# Human-Scene Network: A Novel Baseline with Self-rectifying Loss for Weakly supervised Video Anomaly Detection

Snehashis Majhi[1], Rui Dai[1], Quan Kong[2], Lorenzo Garattoni[3],
Gianpiero Francesca[3], François Brémond[1]
[1] INRIA  [2] Woven Planet Holdings  [3] Toyota Motor Europe

## Abstract

*Video anomaly detection in surveillance systems with only video-level labels (i.e. weakly-supervised) is challenging. This is due to, (i) complex integration of human and scene based anomalies comprising of subtle and sharp spatio-temporal cues in real-world scenarios, (ii) non-optimal optimization between normal and anomaly instances under weak-supervision. In this paper, we propose a Human-Scene Network to learn discriminative representations by capturing both subtle and strong cues in a dissociative manner. In addition, a self-rectifying loss is also proposed that dynamically computes the pseudo temporal-annotations from video-level labels for optimizing the Human-Scene Network effectively. The proposed Human-Scene Network optimized with self-rectifying loss is validated on three publicly available datasets i.e. UCF-Crime, ShanghaiTech and IITB-Corridor, outperforming recently reported state-of-the-art approaches on five out of the six scenarios considered.*

## 1. Introduction

Anomaly detection in real-world videos is a crucial computer vision task thanks to its potential applications in smart cities empowering timely anomaly prevention and investigation. This problem remain unsolved due to the scarcity of spatio-temporally annotated data and the sparsity in occurrence of anomaly events. In consequence, earlier popular methods [1, 3, 6, 8, 8, 11, 12, 17, 20, 26, 29, 30, 34] learn a uni-class (*i.e. only normal class that is easy to acquire*) encoder-decoder network for learning the global temporal regularity and treat abnormality as an out-of-distribution detection (OOD) w.r.t the learned normal distribution. As a matter of fact, these methods fail at learning generalized representations for all possible normal scenarios and hence cause false alarm for unseen normal ones. In light of superior generalization capabilities and detection performance, weakly-supervised video anomaly detection (WS-VAD) methods [19, 27, 35] have recently gained popularity. These methods learn from both normal and anomaly distributions to optimize the separability among the classes with only video-level labels.

Despite the prosperity in mainstream WSVAD ap-

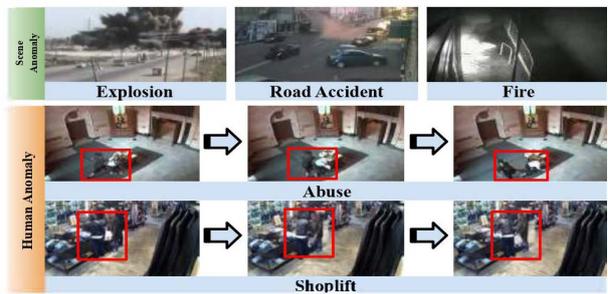

Figure 1. Visualization of scene and human centric anomalies in CCTV videos. The scene-centric anomalies (*row-1*) contain sharp changes of spatio-temporal cues. But, human-centric anomalies may carry strong local motion (*row-2: abuse*) or characterized by subtle features (*row-3: shoplift*).

proaches [4, 10, 19, 27, 31, 32, 35, 36], their performance is still limited due two major challenges: (a) complex real-world abnormalities: difficulties in obtaining discriminative spatio-temporal representations with only video-level labels, when there exists an integration of local human-centric anomalies (*Abuse, Shoplifting, Stealing, Robbery, Vandalism*) with the global scene-centric anomalies (*Explosion, Road Accidents, Fire, Burglary*), as illustrated in Figure 1; (b) non-optimal separation between normal and anomaly classes: obscure consideration of normal and anomaly instances (or temporal segments) in a long untrimmed video labeled as "anomaly" for optimizing the separation among the classes leads to non-optimality.

To address above challenges, attempts have been made in earlier works [19, 27] which first, extract features using a 3D ConvNet and then learn a ranking model by multiple instance learning (MIL) based optimization. Since many previous methods consider only global feature representation (*i.e. features from whole frame*) for optimizing the ranking model, they still lag in detecting human-centric anomalies. This is majorly due to the different features characterizing the anomalies. Scene-centric anomalies are characterized by strong appearance and global motion cues, where as human-centric anomalies have rather subtle and local motion patterns. As a result, global features fail to capture the human-centric subtle cues although they succeed in characterizing scene-centric cues. Furthermore, for optimizing the ranking model, earlier WSVAD approaches adapt a classical MIL loss function proposed in [19] which selects two instances based on the presence of abnormality (*i.e. one each from normal and anomaly videos*) to take part in the optimization process. We believe such an optimization can perform well when anomalies are short in duration like explosion and road accidents, but it may fail drastically when anomalies last longer like shoplifting.

In contrast, we propose a novel Human-Scene Network (HSN) comprising two decoupled sub-networks (subNet) *i.e scene and human subNets*, which are optimized independently with a soft-selection module as the key component for addressing the stringent requirement of WSVAD. The decoupled design of the network helps in learning discriminative representations for scene and human-centric anomalies in a mutually exclusive manner, thus enabling each subNet to detect either coarse scene-centric or fine-grained human-centric anomalies effectively. Now, instead of treating both subNets as independent decision modules to detect anomalies, we propose a class-agnostic soft-selection coupler to choose between the scene and human-centric subNets for a given video. The key difference between the previous WSVAD methods and our approach is outlined in Figure 2. Since the overall performance of the proposed network can be affected due to the limitations of classical MIL loss, we propose a self-rectifying loss function for enhanced optimization with video-level labels. Unlike the earlier loss, the proposed loss not only ensures video-level context maximization between normal and anomaly classes, but also performs an instance-level maximization by dynamically choosing the optimal number of instances empowered by a self error-minimization procedure. The main contributions of this work are as follows:

- A novel Human-Scene Network (HSN) is proposed as a baseline framework for effectively detecting scene and human centric anomalies under weak-supervision.

- A new self-rectifying loss function is proposed for ensuring superior separation between normal and anomaly instances by overcoming the drawbacks of the previous MIL loss.

- An exhaustive experimental analysis is performed to corroborate the robustness of HSN along with the self-rectifying loss on three competitive datasets UCF-Crime [19], ShanghaiTech [13] and IITB-Corridor [18] datasets, outperforming previous approaches on five out of the six scenarios considered.

## 2. Related Work

Weakly-supervised anomaly detection has been studied extensively in the past few years [4, 10, 19, 24, 27, 31, 32, 35, 36]. Majority of previous works follows multiple instance learning (MIL) [16] approach introduced by Sultani

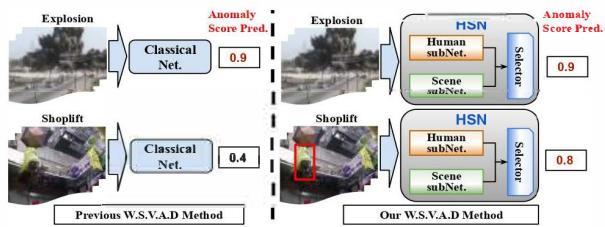

Figure 2. Comparison of previous WSVAD methods with ours. Previous methods considers only scene-centric features (*i.e. from whole frame*) and it fails to detect complex human-centric anomalies. In contrast, our HSN can learn from both cases and obtain better performance.

*et al.* [19] to overcome the drawbacks of traditional unsupervised one-class learning based anomaly detection methods [1, 3, 8, 8, 11, 12, 17, 20, 34]. In weakly-supervised anomaly detection task only video-level labels are provided for learning, authors in [19] only extract off-the-shelf global scene features from a pre-trained 3D ConvNet backbone [2, 23, 28] and aim at training a classification network through a classical ranking loss function. Although superior separation between normal and anomaly instances is ensured by the ranking loss than that of unsupervised anomaly detection methods by choosing only the maximum scoring segment of both normal and anomaly videos for optimization but Sultani *et al.* [19] were able to produce limited detection performance. This is due to, they only focus on global feature representation ignoring the local features and temporal context modeling of videos in order to discriminate anomaly segments at the feature level.

To address temporal context modeling, authors in [32] utilize TCN [9] in MIL based approach to learn temporal dependency encoding for anomaly instances at the feature level. In addition, they also adopt an inner-bag ranking loss function for improved optimization. Another approach [36] combines global optical flow features obtained from PWCNet [21] with the RGB feature map for discriminating the anomaly instances that exhibit strong motion. Similarly, authors in [27] claim that combining audio features obtained from VGGish [7] with global RGB map can discriminate the anomaly instances effectively at the feature level. In addition, Wu *et al.* [27] also utilize Graph Convolution Network (GCN) for temporal context learning in videos leading to improve anomaly detection performance. Recently, Tian *et al.* [22] propose a global temporal feature magnitude based learning paradigm for better separability between maximum and minimum magnitude temporal segments. However, the feature magnitudes are influenced by only strong spatio-temporal variation across temporal segments leading ineffective separability for subtle and local anomalies. Two key drawbacks observed in the above approaches are: (i) consideration of less separable global features to capture real-world anomalies, (ii) optimization with classical ranking loss which only consider selected instances for optimization do not ensure optimal separability.

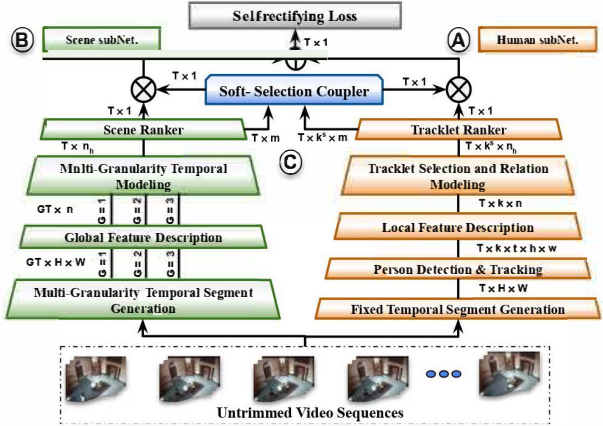

Figure 3. **Human-Scene Network (HSN):** It comprises of three key biulding blocks *i.e.* (A) Human subNet, (B) Scene subNet and (C) Soft-Selection Coupler. HSN inputs a long video sequence that is processed by (A) and (B) to learn either human or scene centric representations synchronously. Afterwards, (C) is optimized to compute a coupled selection factor based on the representation learned by (A) and (B) to focus on either of them.

To combat the limitation of classical ranking loss in obtaining discriminative features, Zhong *et al.* [35] and Feng *et al.* [5] aim at training the 3D ConvNet backbone by generating pseudo temporal annotations through a cleaning noisy labels approach. In this Zhong *et al.* captures the the temporal consistency in a GCN and Feng *et al.* uses a deep MIL ranker module to generate pseudo temporal annotations. Although they obtain higher performance than MIL approaches but still limited by learning the global features in 3D ConvNet and in the pseudo annotation generator. A drawbak lies in these approach is, they operate in two-step manner, where first step is to generate pseudo labels followed by 3D ConvNet optimization. Again as noisy pseudo labels can be result of non-discriminative global features, it can mislead the optimization of 3D ConvNet backbone. Thus, we propose a novel method: Human Scene Network (HSN), which considers both local and global spatio-temporal cues for obtaining discriminative representation in real-world scenarios. In addition we also propose a self-rectifying loss to optimize the HSN which generates pseudo labels to select optimal number of instances in MIL optimization paradigm for maximum separability.

## 3. Human-Scene Network (HSN)

The overview of the proposed HSN is delineated in Figure 3. Its three key components, *human subNet, scene subNet and soft-selection coupler* are designed to precisely detect real-world anomalies when an untrimmed video is given as input. A detailed description of each component in HSN along with the optimization strategy is given in the following subsections.

### 3.1. Scene-subNet

The objective of the Scene-subNet (SsN) is to learn discriminative global representations characterizing scene-centric anomalies. The SsN ensures to capture strong appearance and global motion cues with its four salient building blocks as elaborated below.

**Multi-Granularity Temporal Segment Generation**  Primarily, this section considers a new temporal segment generation strategy ideal for WSVAD. A key drawback in earlier works [19, 22] is, they generate fixed scale (namely granular) temporal segments (*say* $T$) for each video $V$. By dividing $V$ of length $T_l$ into $T$, where $T \ll T_l$, it loses out the fine temporal cues. For this, the anomalies those are short in duration, get suppressed by the neighbouring normal ones and hence lags in detection performance. To preserve fine temporal cues, our strategy divides each $V$ at multiple granularities $G = 1, 2, 3$ for generating temporal segments of size $T$, $2T$ and $3T$ respectively. This temporal segment generation is performed by varying the sampling rate of $V$.

**Global Feature Description**  For global feature description, *off-the-shelf* spatio-temporal features are extracted from a pre-trained 3D ConvNet for each temporal segment ($GT_i$) generated from $V$. A $GT_i$ is a set of consecutive frames that has both background and foreground scene. For a given $GT_i$ the 3D ConvNet extracts a feature map of dimension $c \times n$, where $c$ is the number of 64-frames clips inside $GT_i$ and $n$ is the channel size. Since multiple 64-frame clips can be present inside a $GT_i$, a `max-pooling` is done over $c$ to get a uni feature map per $GT_i$. So for a given $V$ containing $GT$ segments, a feature map $F_{Sc}^{GT}$ of dimension $GT \times n$ is obtained from this block, where $G = 1, 2, 3$.

**Multi-Granularity Temporal Modeling**  To obtain discriminative video representation for long and short anomalies, it is crucial to encode the contextual and fine-grained temporal dependencies respectively. This is due to short and long anomalies are characterized by sharp and progressive change in spatio-temporal features respectively. Hence, high-granular feature map $F_{Sc}^{3T} \in \mathbb{R}^{3T \times n}$ is desirable for short anomalies as it can preserve fine temporal details to capture sharp changes. Similarly, lower-granular feature map $F^T \in \mathbb{R}^{T \times n}$ can succeed in providing the context for long anomalies. So to effectively learn the representations for both long and short anomalies from the pre-computed feature maps $F_{Sc}^T$, $F_{Sc}^{2T}$, $F_{Sc}^{3T}$, a multi-granularity temporal modeling (MGTM) block is proposed in SsN as shown in Figure 4.

MGTM aims at building a temporal feature pyramid followed by dependency modeling by its three modules: (i) temporal downscaler (ii) bottleneck layer and (iii) LSTM cell to encode the contextual and fine-grained temporal dependencies. MGTM obtains the temporal feature pyramid by aggregating the feature maps from multiple temporal granularities *i.e.* $G = 1, 2, 3$. The aggregation is done by first reducing the temporal dimension of higher granular feature map (say $F_{Sc}^{3T}$) and then combining with the successive lower granular feature map (say $F_{Sc}^{2T}$). The temporal downscaler block ensures to reduce the temporal dimension by a factor of $T$ with its two sequential 1D convolution layers with kernel size $k \in \{5, 3\}$ and dilation rate $d \in \{4, 8\}$ respectively. The temporal downscaling is performed by encapsulating the features from neighboring temporal segments with the increased receptive field of 1D dilated convolution layers. The MGTM has two-levels of temporal downscaler *i.e.* level-1 between $G = 3$ and $2$, level-2 between $G = 2$ and $1$. Further, to combine the temporally down scaled feature with lower granular features $F_{Sc}^{GT}$ (where $G = 1, 2$), first $F_{Sc}^{GT}$ is applied to a bottleneck module having one layer of 1D convolution with with kernel size $k = 1$ and then a `concatenate` operation is performed between the feature maps. At the end of final `concatenate` operation that combines the features from $G = 2$ and $1$, it results in a feature map of dimension $T \times 2n_c$ (where $T$= temporal dimension similar to $G = 1$, $n_c$ = number of convolution filters in final bottleneck and temporal downscaler module). The resultant feature map is subsequently input to a `many-to-many` LSTM cell with $n_h$ hidden neurons for global temporal dependency encoding ($F_{Sc}^*$). Since the obtained temporal encodings are enriched by fine-grained and contextual temporal dependency modeling at multiple granularities, it ensures a better discriminative representation among the temporal segments $T_i$.

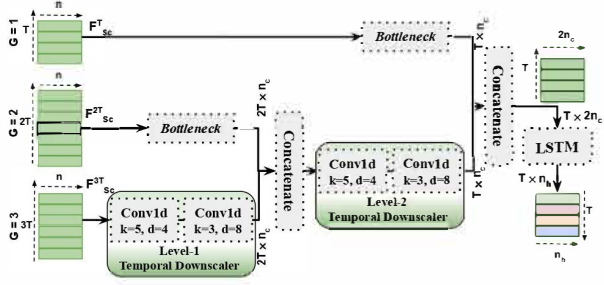

Figure 4. **Multi-Granularity Temporal Modeling (MGTM):** It creates a temporal feature pyramid with *Temporal Downscaler* and *Bottleneck* module followed by a *LSTM* cell to learn discriminative fine-grained and contextual temporal representation.

**Scene Ranker** The scene ranker is a multi-layer perceptron (MLP) with three `fully-connected` (FC) layers which inputs the $F_{Sc}^* \in \mathbb{R}^{T \times n_h}$ to assign anomaly ranks (*or* scores) to each temporal segment. For this, final layer of MLP has a single neuron with *sigmoid* activation to rank each temporal segment independently. Finally the scene ranker outputs a detection score map $D_{Sc}$ of dimension $T \times 1$ to be used in anomaly detection.

### 3.2. Human-subNet

The objective of the Human-subNet (HsN) is to learn discriminative local representation characterising human-centric anomalies. In real-world situations human-centric anomalies can either contain subtle or sharp appearance and motion cues. The HsN ensures to learn a local discriminative representation in all possible scenarios with its four major building blocks as described below.

**Fixed Temporal Segment Generation** Unlike SsN, here the video $V$ is divided into a fixed number of temporal segments, (*say T*). This is because in the following blocks of HsN, humans are detected and tracked in each temporal segment. So, humans are tracked as long as possible making the multi-temporal granularities non-suitable. For instance, with a high granularity value (such as $G = 3$), the length of a segment $T_i$ is reduced by a multiplication factor of $G$, hence the tracking may be lost in such reduced temporal duration. Thus, for a given $V$, HsN outputs $T$ temporal segments of spatial resolution $H \times W$ which are provided to the subsequent block.

**Human Detection and Tracking** In order to learn a local representation in HsN a pre-requisite is to obtain human bounding boxes (BBox) and their corresponding trajectories (*namely tracks*) with a sufficient quality. For this, a pre-trained human detector and tracker network (Byte-Track [33]) is used to extract humans BBox and the tracks in each segment $T_i$. So, for $T$ segments generated from video $V$ with $k$ humans present in the scene, this block outputs a tracklet map of dimension $T \times k \times t \times h \times w$, where $t$ is the maximum track duration of each tracklet and $h \times w$ is the BBox dimension.

**Local Feature Description** A pre-trained 3D ConvNet is used to sequentially extract spatio-temporal features for each tracklet $k_j$ present in each temporal segment $T_i$. Since the 3DConvNet used in HsN is bounded by the input size of 64 frames, each with resolution $224 \times 224$, the tracklet map is resized and padded to meet the input requirements. For a given tracklet $k_j$ present in $T_i$, the track duration $t$ can be larger than 64-frame sequences. So, a `max-pooling` operation is performed to obtain a $n$-dimensional unitary feature vector per tracklet $k_j$ in each $T_i$. The obtained local feature map $F_{Tr} \in \mathbb{R}^{T \times k \times n}$ is fed to the subsequent block to encode the dependency among tracklets.

**Tracklet Selection and Relation Modeling** A key intricacy of encoding the individual tracklets behaviour and relation among them lies in number of people present in the video. With a large number of people, the relation modeling seems difficult and increases the model complexity as well. For this, a tracklet selection (TS) strategy targeted only for anomaly detection task is proposed here to filter out the salient tracklets, followed by a relation modeling method to obtain local discriminative representation in an effective and efficient manner.

The TS filters out $k^s$ salient people out of $k$ in $F_{Tr}$, where $k^s$ people has significantly distinctive behaviour (*assuming sharp change in the spatio-temporal feature space as distinctive*) than that of remaining (*i.e.* $k - k^s$). For a given spatio-temporal feature vector $X$ corresponding to tracklet $k_j$, the distinctive behaviour is defined by computing the feature magnitude (FM) of $X$ that captures the variation of appearance and motion cues in the spatio-temporal feature space. Formally, $FM(X) = \sum \|X\|_2$. So, the proposed selection strategy starts by computing $FM$ for all $k$ followed by arranging them in descending order and then keeps top $k^s$ to be used in relation modeling, as computed by $TS(F_{Tr}) = \max(\sum_{j=1}^{j=k} \|F_{Trj}\|_2)$. With the top $k^s$ tracklets where $FM(k^1) \leq FM(k^2) \leq \ldots \leq FM(k^s)$, a LSTM cell with $n_h$ hidden neurons is used to output a fixed order dependency encoded feature map $F_{Tr}^*$ the as a relation modeling among the selected tracklets.

**Tracklet Ranker** The tracklet ranker is identical to the scene ranker which inputs $F_{Tr}^* \in \mathbb{R}^{T \times k^s \times n}$ to assign anomaly scores to each tracklet present in temporal segments. The difference lies in obtaining the temporal detection score map $D_{Tr}$ of dimension $T \times 1$ which is computed by applying a `max-pooling` operator over $k^s$.

### 3.3. Soft-Selection Coupler

In order to detect effectively both human and scene centric anomalies by selecting either HsN or SsN, a Soft selection coupler (SSC) is proposed in HSN. The SSC shown in Figure 5 computes two selection factors *i.e.* $S_{HsN}$ and $S_{SsN}$ by inputting the intermediate representations of tracklet ranker $F_T \in \mathbb{R}^{T \times k^s \times m}$ and Scene ranker $F_S \in \mathbb{R}^{T \times m}$, where $T$ is the number of temporal segments, $k^s$ is the number of tracklets and $m$ is the channel size. Essentially, SSC has three blocks: (i) segment-level, (ii) video-level and (iii) final selection block to output $S_{HsN}$ and $S_{SsN}$. The *segment-level selection* block computes two attention weights *i.e.* $A_{HsN}^S \in \mathbb{R}^{T \times 1}$ and $A_{SsN}^S \in \mathbb{R}^{T \times 1}$ for each temporal segment $T_i$ signifying the weighted association of each $T_i$ to HsN and SsN. In this block, a `max-pooling` operator is first applied over $k^s$ dimension of $F_T$ to identically match with $F_S$ dimension. The output of max-pooling operation is denoted by $F_T^M$. Then, $F_T^M$ and $F_S$ are projected to two parallel FC layers followed by *ReLu* activation for latent space representation. This dissociative latent representations are coupled by a `concatenate` operator. The coupled representation is then applied to two parallel `sigmoid` activated FC layers, each having one units to compute $A_{HsN}^S$ and $A_{SsN}^S$ in a mutually exclusive manner. The segment-level selection block outputs $A_{HsN}^S$ and $A_{SsN}^S$ by learning fine-grained temporal representation of a video.

Dissimilar to this, *video-level selection* block encodes the association of a video by computing two attention weights *i.e.* $A_{HsN}^V \in \mathbb{R}^{1 \times 1}$ and $A_{SsN}^V \in \mathbb{R}^{1 \times 1}$ from the contextual temporal representation. For this, temporal contextual initialization is done by applying a `average-pooling` operation over $T$ to the $F_T^M$ and $F_S$ feature map. Then the contextual representations are entangled by a `concatenate` operation. This feature map is then projected to two parallel `sigmoid` activated FC layers, each having one units to compute $A_{HsN}^V$ and $A_{SsN}^V$ in a mutually exclusive manner. The output from segment-level and video-level selection blocks are masked in *final selection* block to obtain $S_{HsN} \in \mathbb{R}^{T \times 1}$ and $S_{SsN} \in \mathbb{R}^{T \times 1}$. For masking, the $A_{HsN}^V$ and $A_{SsN}^V$ are first *inflated* across $T$ to match the dimension of $A_{HsN}^S$ and $A_{SsN}^S$ and then the corresponding weights are combined by a *Hadamard product*.

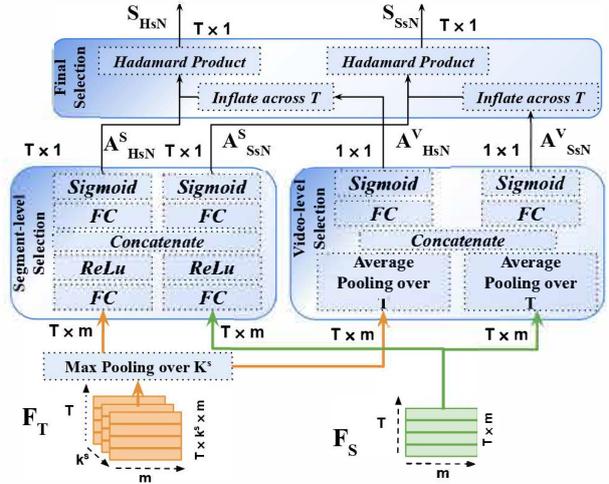

Figure 5. **Soft-Selection Coupler (SSC):** It learns two selection factors $S_{HsN}$ and $S_{SsN}$ to couple the local and global representations learned by HsN and SsN respectively.

**Coupled Detection Score:** The final detection score ($D$) of HSN is computed by coupling the selection factors $S_{HsN}$ and $S_{SsN}$ with the human and scene subNet detection scores *i.e* $D_{Tr}$ and $D_{Sc}$ respectively. The coupling is done by multiplying the selection factors with the corresponding detection scores and then the results are added to obtain $D$. Formally, $D = (S_{HsN} \times D_{Tr}) + (S_{SsN} \times D_{Sc})$.

### 3.4. Optimization of Human-Scene Network

The proposed human-scene network (HSN) is end-to-end trainable (excluding the local and global feature extractors) with a novel self-rectifying loss. Although HSN follows multiple instance learning (MIL) optimization paradigm similar to earlier WSVAD works [19], the error computation procedure differs significantly. In this, due to the unavailability of precise temporal annotation for each video, the error is computed by considering two bags of instances, namely $D_a$ and $D_n$. $D_a$ and $D_n$ are collection of detection scores ($D$) corresponding to the temporal segments extracted from anomaly and normal video sequences respectively.

**Self-rectifying Loss:** It is designed in such a way that, it not only ensures context level but also performs instance level score maximization by generating a pseudo temporal annotation for each segment. The aim of pseudo label generation is to choose the correct number of anomaly and normal instances to take part in the optimization. The pseudo temporal annotations are computed during each iteration of optimization (*i.e. one step*) to avoid two-step optimization paradigm [5, 35] (*where, first step to generate pseudo annotations, second step for separability maximization*). Since our loss function is based on one-step optimization and since the pseudo annotations computed at the initial iteration can be noisy, a *self-rectification* mechanism is also proposed to refine the labels during the optimization. The proposed self-rectifying loss is presented in eq (1), which is a weighted sum of two components $L_C$ and $L_I$ to perform context and instance level maximization respectively.

$L_C$ first computes the sum of scores from all $T$ instances present in $D_a$ and $D_n$ to obtain video context and then maximize the separation among them to ensure context separability. With only $L_C$, it is not sufficient to ensure optimal separability between normal and anomaly classes, since video containing short anomalies will be suppressed by neighboring normal scores and hence can not be optimized effectively.

$$L_{SR}(D_a, D_n) = \underbrace{\lambda_1 \max(0, 1 - \sum_{i=1}^{T}(D_a^i) + \sum_{i=1}^{T}(D_n^i))}_{L_C}$$
$$+ \underbrace{\lambda_2 \|Err(Correct) - Err(Noisy)\|}_{L_I} \quad (1)$$

$$Err(X) = \begin{cases} \underbrace{\frac{1}{T}\sum_{i=1}^{T}(D_n^i - P_{y_n}^i)^2,}_{MSE(D_n)} & \text{if } X = Correct \\ \forall i, \quad P_{y_n}^i = 0 \\ \underbrace{\frac{1}{T}\sum_{i=1}^{T}(D_a^i - P_{y_a}^i)^2,}_{MSE(D_a)} & \text{if } X = Noisy \\ \forall i, \text{ if } D_a^i \leq D_{ref} \text{ then } P_{y_a}^i = 0, \\ \forall i, \text{ if } D_a^i > D_{ref} \text{ then } P_{y_a}^i = 1 \end{cases}$$
(2)

For this, $L_I$ performs the instance-level optimization in $D_a$ and $D_n$ by generating pseudo-temporal labels which are empowered by a self-rectification mechanism. Since $D_n$ contains no anomaly instances, so the pseudo temporal label for segment $i$ ($Py_n^i$) for $D_n$ is correct (*i.e. $\forall i, Py_n^i = 0$*[1]). In contrast, $D_a$ contains a mixed distribution (*i.e. normal and anomaly*) with no prior knowledge, so the pseudo temporal labels for $D_a$ (*i.e. $Py_a^i$*) are noisy. In $D_a$, the pseudo temporal labels $Py_a^i$ are computed by comparing their prediction scores ($D_a^i$) to a dynamic reference point ($D_{ref}$). The $D_{ref}$ for $D_a$ is calculated by taking the average between the maximum and minimum score of $D_a$, $D_{ref} = (\max(D_a) + \min(D_a))/2$, which essentially provides a dynamic threshold point for each video characterizing different categories of anomaly. The quality of $Py_a$ computation depends on choosing an appropriate $D_{ref}$, which may be a rough approximation at the initial iteration. Hence, for optimal instance-level separation, it is necessary to rectify the noisy pseudo temporal labels (*i.e. $Py_a$*) with the help of the correct ones (*i.e. $Py_n$*). So, to rectify the noisy distribution, $L_I$ minimizes the difference between the errors ($Err$) obtained from the correct pseudo labels to the noisy pseudo labels. It can be observed that $Err(Noisy)$ has a direct dependency on the $D_{ref}$ point. So, by minimizing $\|Err(Correct) - Err(Noisy)\|$, the $D_{ref}$ gets adjusted and reaches an optimal point where $Py_a$ has minimum noise. Thus, this self-rectification procedure enables the noisy pseudo labels to get rectified with the guidance of correct labels. Kindly note, for computing $Err$, we adopt mean-squared-error (MSE) between the predicted scores and their corresponding pseudo labels as portrayed in eq (2).

---
[1] Kindly note, we refer label = 0 for normal and label = 1 for anomaly instances.

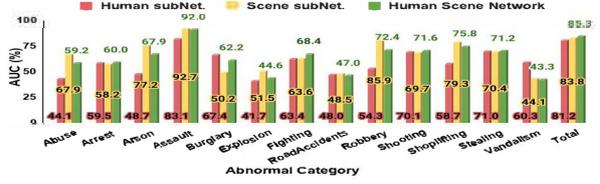

Figure 6. Category wise detection performance (AUC%) for UCF-Crime dataset quantifying effectiveness of HSN.

## 4. Experiments

### 4.1. Datasets and Evaluation Metric

The experiments are conducted on three publicly anomaly detection datasets, namely, UCF-Crime [19], ShanghaiTech [13] and IITB-Corridor [18]. Following [5, 19, 22, 35], frame-level Receiver Operating Characteristics (ROC) and its corresponding Area Under the Curve (AUC) are used to evaluate the anomaly detection performance. For UCF-Crime dataset, category wise detection performance is also computed on both official test split [19] and whole dataset (*i.e. using 5-fold cross validation*) to evaluate the robustness of detection performance in various critical situations. For the 5-fold cross-validation, we report the mean-AUC (mAUC) of all 5-folds to evaluate the method. Since for the 5-fold evaluation, we need the temporal annotation of the complete UCF-Crime dataset, we obtained it from Wan *et al* [25]. **Kindly refer to supplementary material for complete dataset descriptions and implementation details.**

| Human subNet. | | Scene subNet. | | SSC | | AUC(%) |
|---|---|---|---|---|---|---|
| PDT | TSRM | GS | MGTM | SLS | VLS | |
| ✓ | ✗ | ✗ | ✗ | ✗ | ✗ | 76.53 |
| ✓ | ✓ | ✗ | ✗ | ✗ | ✗ | 81.21 |
| ✗ | ✗ | ✓ | ✗ | ✗ | ✗ | 78.62 |
| ✗ | ✗ | ✓ | ✓ | ✗ | ✗ | 83.78 |
| ✓ | ✓ | ✓ | ✓ | ✓ | ✗ | 84.52 |
| ✓ | ✓ | ✓ | ✓ | ✓ | ✓ | 85.30 |

Table 1. Ablation on each component of HSN on UCF-Crime dataset in terms of AUC(%). Here, SLS:Segment-level selection, VLS:Video-level selection block of SSC.

## 4.2. Ablation Study

A detailed and sequential ablation study is carried out in this section to quantify the two novel contributions *i.e.* human-scene network (HSN) and self-rectifying loss. For all ablation studies, UCF-Crime [19] dataset is chosen as it has a good number of human and scene based anomalies.

**Effectiveness of HSN :** As HSN comprises of multiple building blocks, each block is evaluated in terms of anomaly detection performance as shown in Table 1. At first, human and scene subNets are independently considered for experimentation. The temporal segment generation blocks, feature descriptors (*i.e.* I3D-ResNet50) and ranker block are inherently added to both subNets for determining detection performances. In human subNet, the human baseline experiment is performed with only the tracklets obtained from people detection and tracking (PDT) method. Subsequently by adding tracklet selection and relation modeling (TSRM) block to human subNet, it boosts the performance ($+4.68\%$) significantly compared to the human baseline. Similarly, in scene subNet, the global scene (GS) is only taken at the beginning to define the scene baseline. On top of that, adding multi-granularity temporal modeling (MGTM) block improves the detection performance ($+5.16\%$) by a large margin. The performance boost for both human and scene subNets outlines the potentiality and significance of TSRM and MGTM blocks respectively.

Further, in order to verify the complementary nature of representations learned by both subNets, we visualize the category wise detection performance as in Figure 6. It is shown that for human-centric anomaly categories such as *arrest, burglary, fighting, stealing, vandalism* the performance of human subNet is superior to the one of scene subNet and vice versa for scene-centric anomaly category like *arson, explosion*. The soft selection coupler (SSC) is introduced in HSN to take benefits from both subNets and to improve the overall detection performance. Table 1 shows that with only segment-level selection of SSC (*i.e.* $A^S_{HsN}$ and $A^S_{SsN}$) of SSC, the detection performance is improved by $3.31\%$ and $0.74\%$ compared to individual subNets. Finally by combining both segment-level and video-level selection in SSC (*i.e.* $S_{HsN}$ and $S_{SsN}$), it further boosts the detection performance by $4.07\%$ and $2.5\%$ compared to that of only human and scene subNet respectively. From Figure 6, it can be inferred that thanks to the decoupled design of HSN, distinctive features are learned for human and scene based anomalies and the SSC effectively combine the complementary representations to boost the performance.

| Method | Optimization | | AUC(%) |
|---|---|---|---|
| | $L$ [19] | $L_{SR}$ [eq 1] | |
| Sultani *et al.* [19] | ✓ | ✗ | 77.42 |
| | ✗ | ✓ | 78.39 |
| Majhi *et al.* [15] | ✓ | ✗ | 81.88 |
| | ✗ | ✓ | 83.27 |
| HSN | ✓ | ✗ | 82.76 |
| | ✗ | ✓ | 85.30 |

Table 2. Ablation Study to showcase the AUC gain by $L_{SR}$ loss in various methods evaluated on UCF-Crime dataset.

**Effectiveness of $L_{SR}$ :** In order to gain enhanced performance, optimization with $L_{SR}$ plays a key role in HSN. To corroborate the effectiveness and adaptability of $L_{SR}$, an experimental ablation study is performed as shown in Table 2. We choose two recently reported MIL based methods *i.e.* Sultani *et al.* [19] and Majhi *et al.* [15] for comparison. Those methods were previously optimized with classical ranking loss ($L$). Optimizing the Sultani *et al.* and Majhi *et al.* methods with $L_{SR}$ gives $0.97\%$ and $1.39\%$ performance gain respectively. Similarly, HSN optimized with $L_{SR}$ boosts the detection performance by $2.52\%$ compared to $L$. Since previous methods operate on coarse spatio-temporal features which have lower separabilities compared to HSN representation, so the performance boost by $L_{SR}$ is marginal in those methods. But in HSN, a superior discriminative representation is captured due to the decoupling of scene and human cues and hence it generates less noisy pseudo labels leading to greater separable features.

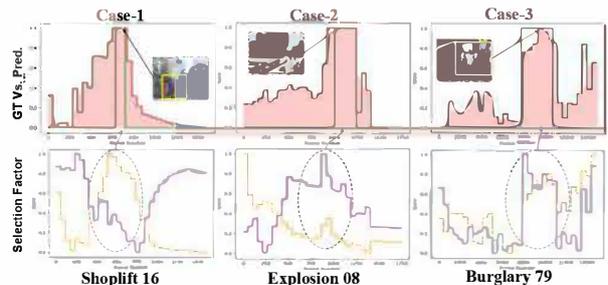

Figure 7. Visualization of Ground truth (Green shed) vs. prediction scores (Red shed) for various cases in Row-1. Row-2 is the selection factors $S_{HsN}$ (Yellow plot) and $S_{SsN}$ (Violet plot) to focus on either HsN or SsN for effective coupling.

## 4.3. Qualitative Analysis

In Figure 7, we show the prediction scores (Row-1) obtained from proposed method along with the selection factors (Row-2) corresponding to human and scene subNets (*i.e.* HsN and SsN) in three cases (**case-1: human centric, case-2: scene centric, case-3: both human and scene**) of anomaly scenarios. The samples "Shoplifting16", "Explosion08", and "Burglary79" give an overview of the performance for all three cases. In "Shoplifting16" where the anomaly is being done by a women, the effective detection score is majorly influenced by HsN since higher selection factor to HsN is assigned to the localized area. Similarly, for "Explosion08" where a bomb blast is recorded, we see the detection performance is triggered due to higher factor assigned to SsN. From both anomaly cases, the proposed method yields superior detection performance by effectively selecting either subNet. Moreover, we choose a third case where the anomaly is characterized by both human and scene localized areas. "Burglary79" consists of such a scenario where a group of criminals are illegally entering to a building by breaking the entrance.

| | Official Test-Evaluation [19] (AUC%). | | | | | | | | | | | | |
|---|---|---|---|---|---|---|---|---|---|---|---|---|---|
| Methods | Abuse | Arrest | Arson | Assault | Burglary | Explosion | Fighting | RoadAcc. | Robbery | Shooting | Shoplifting | Stealing | Vandalism | Total |
| Tian et al. [22] | 55.93 | 59.16 | 65.32 | 70.71 | 70.11 | 45.28 | 70.02 | 55.94 | 69.84 | 73.63 | 70.8 | 75.17 | 64.31 | 84.30 |
| Ours | 59.21 | 60.02 | 67.86 | 91.95 | 62.19 | 44.57 | 68.36 | 47.01 | 72.37 | 71.63 | 75.78 | 71.18 | 43.26 | 85.30 |
| | 5-Fold Evaluation.(mean-AUC%) | | | | | | | | | | | | |
| Tian et al. [22] | 56.3 | 56.6 | 62.9 | 75.1 | 74.2 | 69.8 | 58.3 | 69.9 | 59.2 | 62.0 | 54.1 | 59.2 | 61.9 | 63.3 |
| Ours | 66.7 | 61.8 | 66.0 | 81.1 | 67.2 | 58.8 | 65.6 | 57.9 | 70.5 | 71.4 | 65.0 | 67.0 | 57.6 | 65.6 |

Table 3. Category wise performance comparison with a Tian et al. to showcase the AUC gain by our in various categories of UCF Crime dataset. We made the comparison in both official test split [19] and 5-fold evaluation to justify the robustness of our method.

| Methods | Feature | UCF-C | ST | IITB-C |
|---|---|---|---|---|
| | | | AUC(%) | |
| Sultani et al. [19] | C3D | 75.41 | - | - |
| | I3D-Inc | 77.42 | 80.02 | 74.59 |
| Zhang et al. [32] | C3D | 78.66 | - | - |
| Zhu et al. [36] | C3D | 79.00 | - | - |
| Zhong et al. [35] | C3D | 81.08 | 76.44 | - |
| | TSN-Inc | 82.12 | 84.44 | - |
| Feng et al. [5] | I3D-Inc | 82.30 | - | - |
| Majhi et al. [14] | I3D-Inc | 82.12 | - | - |
| Wu et al. [27] | I3D-Inc | 82.44 | 85.38 | - |
| Majhi et al. [15] | I3D-Inc | 82.67 | 88.86 | 78.07 |
| Tian et al. [22] | I3D-Res | 84.30 | 97.21 | 79.82* |
| Our | I3D-Inc | 84.33 | 93.72 | 80.37 |
| | I3D-Res | 85.30 | 96.22 | 82.16 |

Table 4. State-of-the-art performance comparisons in terms of frame-level AUC on UCF-Crime (UCF-C), ShanghaiTech (ST) and IITB-Corridor (IITB-C) dataset. Kindly Note: * marked AUC is our implementation.

### 4.4. State-of-the-art Comparison

**Overall Performance Comparison** In Table 4, we compare the overall performance of the proposed method with recently reported state-of-the-art methods for UCF-Crime, ShanghaiTech and IITB-Corridor datasets. For a fair comparison, as several feature extractor backbones are used to report AUC in previous methods, we report results using two widely used backbones i.e. Inception-v1 I3D (I3D-Inc) and ResNet50 I3D (I3D-Res) for the three datasets.

In UCF-Crime, our method outperforms the recent I3D-Inc and I3D-Res based method of Majhi et al. [15] and Tian et al. [22] by +1.66% and +1% margin respectively. For ShanghaiTech dataset, although our method outperforms the recent I3D-Inc based method of Majhi et al. [15] by +4.86%, but the method fails to achieve better performance than [22] with I3D-Res backbone. This is due to ShanghaiTech, that has only 65 anomaly videos collected from a small and focused data distribution for training. It contains only simple human anomalies (e.g. run, fall down, ride cycle etc.) characterized by strong motion, so only few events are not correctly detected. Due to the small number of anomaly samples, our HSN could not be optimized sufficiently to outperform methods dedicated to strong motion events. Further, to confirm the robustness of our method in complex and diverse anomaly distributions, we have also validated our method on the IITB-Corridor dataset. In this, it surpasses the recent I3D-Inc method [15] and I3D-Res method [22] by +2.3% and +2.34% margin respectively.

**Category Wise Performance Comparison** Further, we also compare the abnormal category wise performance with Tian et al. [22] in UCF-Crime dataset. First, we made the comparison in the official test split [19] and found that our method is superior in detecting human-centric abnormal categories like *abuse, arrest, arson, assault, robbery, shoplifting* compared to Tian et al. [22] as reported in Table 3. However, the performance gain is not significant. This is due to fewer number of samples in the official test set. For this, we perform the 5-fold cross-validation and report the mean-AUC (mAUC) of 5-folds for categories-wise performance comparison. Since it covers the entire UCF-Crime dataset, we believe it provides a more robust and justified performance w.r.t official test set [19]. From the 5-fold cross-validation, we found that our method outperforms Tian et al. [22] in all fine-grained and subtle human-centric anomalies (*abuse, arrest, arson. assault, fighting, robbery, shooting, shoplifting, stealing*) by a significant margin as reported in Table 3. As a result, our method surpasses Tian et al. [22] by 2.3% margin in total performance in K-fold evaluation. But, our method lies behind Tian et al. on simple scene based anomalies (*explosion, bruglary, road accidents*) and on human anomalies like *vandalism* (humans tampering the cameras) where there exists a sharp change

| Methods | FLOPs(G) | Speed(FPS) |
|---|---|---|
| PDT [33] | 281.9 | 30 |
| Majhi *et al.* [15](I3D-Inc) | 108.1 | 267 |
| Tian *et al.* [22](I3D-Res) | 153.2 | 211 |
| HSN(I3D-Inc) w-o PDT | 108.7 | 227 |
| HSN(I3D-Res) w-o PDT | 153.7 | 159 |

Table 5. Complexity comparison of Human-Scene Network. Here, G: Giga, FPS: Frames-per-second,w-o: with-out.

in the scene.

### 4.5. Network Complexity Analysis

In this section, a complexity analysis of HSN is performed to meet real-world applicability. Since HSN rely on people detection and tracking (PDT) method, the complexity of the PDT method (*ByteTrack* [33]) is evaluated first and then the complexity of anomaly detection methods are reported and compared w.r.t FLOPs, and speed as shown in Table 5. It can be seen that our HSN is computationally competitive in terms of FLOPs and speed w.r.t recently reported articles. For fair comparison, evaluation is done on a single 2080Ti GPU. Considering recent CCTV cameras which operates in 30 FPS, HSN can detect anomalies effectively in near real-time.

## 5. Conclusion

In this work, we presented a novel human scene network (HSN) optimized by a self-rectifying loss function as a baseline to detect real-world anomalies under weak-supervision. The HSN ensures superior detection performance in complex real world scenarios for two reasons: First, the decoupled design of HSN comprising Human and Scene subNet followed by soft-selection coupler can effectively learn a local and global discriminative representations for human and scene centric anomalies, respectively. Second, optimizing HSN with the proposed self-rectifying loss ensures greater separability between the classes. From experimentation, it can be noted that the proposed method gain competitive performance in five out of six scenarios considered compared to the recently reported methods for three popular datasets.

**Limitation-and-Contribution Trade-off :** As HSN rely on people detection and tracking method for learning human centric representation, it induces slightly more complexity compared to the previous methods. However, it opens up new directions to analyse the complex abnormal scenarios in a more fine-grained way to address the real-world challenges which is majorly missing in earlier weakly-supervised methods. From the performance and complexity analysis prospective, our method establish a good trade off to meet the real-world applicability.

## References

[1] Amit Adam, Ehud Rivlin, Ilan Shimshoni, and Daviv Reinitz. Robust real-time unusual event detection using multiple fixed-location monitors. *IEEE transactions on pattern analysis and machine intelligence*, 30(3):555–560, 2008. 1, 2

[2] Joao Carreira and Andrew Zisserman. Quo vadis, action recognition? a new model and the kinetics dataset. In *The IEEE Conference on Computer Vision and Pattern Recognition (CVPR)*, July 2017. 2

[3] Yang Cong, Junsong Yuan, and Ji Liu. Abnormal event detection in crowded scenes using sparse representation. *Pattern Recognition*, 46(7):1851–1864, 2013. 1, 2

[4] Shikha Dubey, Abhijeet Boragule, and Moongu Jeon. 3d resnet with ranking loss function for abnormal activity detection in videos. *arXiv preprint arXiv:2002.01132*, 2020. 1, 2

[5] Jia-Chang Feng, Fa-Ting Hong, and Wei-Shi Zheng. Mist: Multiple instance self-training framework for video anomaly detection. In *Proceedings of the IEEE/CVF Conference on Computer Vision and Pattern Recognition*, pages 14009–14018, 2021. 3, 6, 8

[6] Mahmudul Hasan, Jonghyun Choi, Jan Neumann, Amit K. Roy-Chowdhury, and Larry S. Davis. Learning temporal regularity in video sequences. In *The IEEE Conference on Computer Vision and Pattern Recognition (CVPR)*, June 2016. 1

[7] Shawn Hershey, Sourish Chaudhuri, Daniel PW Ellis, Jort F Gemmeke, Aren Jansen, R Channing Moore, Manoj Plakal, Devin Platt, Rif A Saurous, Bryan Seybold, et al. Cnn architectures for large-scale audio classification. In *2017 ieee international conference on acoustics, speech and signal processing (icassp)*, pages 131–135. IEEE, 2017. 2

[8] Jaechul Kim and Kristen Grauman. Observe locally, infer globally: a space-time mrf for detecting abnormal activities with incremental updates. In *2009 IEEE Conference on Computer Vision and Pattern Recognition*, pages 2921–2928. IEEE, 2009. 1, 2

[9] Colin Lea, Rene Vidal, Austin Reiter, and Gregory D Hager. Temporal convolutional networks: A unified approach to action segmentation. In *European conference on computer vision*, pages 47–54. Springer, 2016. 2

[10] Shuheng Lin, Hua Yang, Xianchao Tang, Tianqi Shi, and Lin Chen. Social mil: Interaction-aware for crowd anomaly detection. In *2019 16th IEEE International Conference on Advanced Video and Signal Based Surveillance (AVSS)*, pages 1–8. IEEE, 2019. 1, 2

[11] Wen Liu, Weixin Luo, Dongze Lian, and Shenghua Gao. Future frame prediction for anomaly detection–a new baseline. In *Proceedings of the IEEE conference on computer vision and pattern recognition*, pages 6536–6545, 2018. 1, 2

[12] Cewu Lu, Jianping Shi, and Jiaya Jia. Abnormal event detection at 150 fps in matlab. In *Proceedings of the IEEE international conference on computer vision*, pages 2720–2727, 2013. 1, 2

[13] Cewu Lu, Jianping Shi, and Jiaya Jia. Abnormal event detection at 150 fps in matlab. In *Proceedings of the IEEE international conference on computer vision*, pages 2720–2727, 2013. 2, 6

[14] Snehashis Majhi, Srijan Das, Francois Bremond, Ratnakar Dash, and Pankaj Kumar Sa. Weakly-supervised

joint anomaly detection and classification. *arXiv preprint arXiv:2108.08996*, 2021. 8

[15] Snehashis Majhi, Srijan Das, and François Brémond. Dam: Dissimilarity attention module for weakly-supervised video anomaly detection. In *2021 17th IEEE International Conference on Advanced Video and Signal Based Surveillance (AVSS)*, pages 1–8, 2021. 7, 8, 9

[16] Oded Maron and Tomás Lozano-Pérez. A framework for multiple-instance learning. In *Proceedings of the 1997 Conference on Advances in Neural Information Processing Systems 10*, NIPS '97, page 570–576, Cambridge, MA, USA, 1998. MIT Press. 2

[17] Bharathkumar Ramachandra and Michael Jones. Street scene: A new dataset and evaluation protocol for video anomaly detection. In *The IEEE Winter Conference on Applications of Computer Vision*, pages 2569–2578, 2020. 1, 2

[18] Royston Rodrigues, Neha Bhargava, Rajbabu Velmurugan, and Subhasis Chaudhuri. Multi-timescale trajectory prediction for abnormal human activity detection. In *The IEEE Winter Conference on Applications of Computer Vision (WACV)*, March 2020. 2, 6

[19] Waqas Sultani, Chen Chen, and Mubarak Shah. Real-world anomaly detection in surveillance videos. In *Proceedings of the IEEE Conference on Computer Vision and Pattern Recognition*, pages 6479-6488, 2018. 1, 2, 3, 5, 6, 7, 8

[20] Che Sun, Yunde Jia, Yao Hu, and Yuwei Wu. Scene-aware context reasoning for unsupervised abnormal event detection in videos. In *Proceedings of the 28th ACM International Conference on Multimedia*, pages 184–192, 2020. 1, 2

[21] Deqing Sun, Xiaodong Yang, Ming-Yu Liu, and Jan Kautz. Pwc-net: Cnns for optical flow using pyramid, warping, and cost volume. In *Proceedings of the IEEE Conference on Computer Vision and Pattern Recognition*, pages 8934–8943, 2018. 2

[22] Yu Tian, Guansong Pang, Yuanhong Chen, Rajvinder Singh, Johan W Verjans, and Gustavo Carneiro. Weakly-supervised video anomaly detection with robust temporal feature magnitude learning. In *Proceedings of the IEEE/CVF International Conference on Computer Vision*, pages 4975–4986, 2021. 2, 3, 6, 8, 9

[23] Du Tran, Lubomir Bourdev, Rob Fergus, Lorenzo Torresani, and Manohar Paluri. Learning spatiotemporal features with 3d convolutional networks. In *The IEEE International Conference on Computer Vision (ICCV)*, December 2015. 2

[24] Boyang Wan, Yuming Fang, Xue Xia, and Jiajie Mei. Weakly supervised video anomaly detection via center-guided discriminative learning. In *2020 IEEE International Conference on Multimedia and Expo (ICME)*, pages 1–6. IEEE, 2020. 2

[25] Boyang Wan, Wenhui Jiang, Yuming Fang, Zhiyuan Luo, and Guanqun Ding. Anomaly detection in video sequences: A benchmark and computational model. *IET Image Processing*, 15(14):3454–3465, 2021. 6

[26] Jue Wang and Anoop Cherian. Gods: Generalized one-class discriminative subspaces for anomaly detection. In *Proceedings of the IEEE International Conference on Computer Vision*, pages 8201–8211, 2019. 1

[27] Peng Wu, Jing Liu, Yujia Shi, Yujia Sun, Fangtao Shao, Zhaoyang Wu, and Zhiwei Yang. Not only look, but also listen: Learning multimodal violence detection under weak supervision. In *European Conference on Computer Vision*, pages 322–339. Springer, 2020. 1, 2, 8

[28] Huijuan Xu, Abir Das, and Kate Saenko. R-c3d: Region convolutional 3d network for temporal activity detection. In *Proceedings of the IEEE international conference on computer vision*, pages 5783–5792, 2017. 2

[29] Jongmin Yu, Younkwan Lee, Kin Choong Yow, Moongu Jeon, and Witold Pedrycz. Abnormal event detection and localization via adversarial event prediction. *IEEE Transactions on Neural Networks and Learning Systems*, 2021. 1

[30] M Zaigham Zaheer, Arif Mahmood, M Haris Khan, Mattia Segu, Fisher Yu, and Seung-Ik Lee. Generative cooperative learning for unsupervised video anomaly detection. In *Proceedings of the IEEE/CVF Conference on Computer Vision and Pattern Recognition*, pages 14744–14754, 2022. 1

[31] Muhammad Zaigham Zaheer, Arif Mahmood, Hochul Shin, and Seung-Ik Lee. A self-reasoning framework for anomaly detection using video-level labels. *IEEE Signal Processing Letters*, 27:1705–1709, 2020. 1, 2

[32] Jiangong Zhang, Laiyun Qing, and Jun Miao. Temporal convolutional network with complementary inner bag loss for weakly supervised anomaly detection. In *2019 IEEE International Conference on Image Processing (ICIP)*, pages 4030-4034. IEEE, 2019. 1, 2, 8

[33] Yifu Zhang, Peize Sun, Yi Jiang, Dongdong Yu, Zehuan Yuan, Ping Luo, Wenyu Liu, and Xinggang Wang. Bytetrack: Multi-object tracking by associating every detection box. *arXiv preprint arXiv:2110.06864*, 2021. 4, 9

[34] B. Zhao, L. Fei-Fei, and E. P. Xing. Online detection of unusual events in videos via dynamic sparse coding. In *CVPR 2011*, pages 3313–3320, June 2011. 1, 2

[35] Jia-Xing Zhong, Nannan Li, Weijie Kong, Shan Liu, Thomas H. Li, and Ge Li. Graph convolutional label noise cleaner: Train a plug-and-play action classifier for anomaly detection. In *The IEEE Conference on Computer Vision and Pattern Recognition (CVPR)*, June 2019. 1, 2, 3, 6, 8

[36] Yi Zhu and Shawn Newsam. Motion-aware feature for improved video anomaly detection. *arXiv preprint arXiv:1907.10211*, 2019. 1, 2, 8